\documentclass[runningheads]{llncs}

\usepackage{eccv}

\usepackage{eccvabbrv}

\usepackage{graphicx}
\usepackage{booktabs}

\usepackage[accsupp]{axessibility}  

\usepackage{wrapfig}
\usepackage{graphicx}
\usepackage{amsmath}
\usepackage{amssymb}
\usepackage{booktabs}
\usepackage{dsfont}
\usepackage{makecell}
\usepackage{booktabs}
\usepackage[dvipsnames]{xcolor}
\usepackage{arydshln}
\usepackage{pifont}
\usepackage{xcolor}
\usepackage{array}
\usepackage{colortbl}

\usepackage{multirow}

\usepackage{hyperref}

\usepackage{orcidlink}

\begin{document}

\title{Revisit Event Generation Model: Self-Supervised Learning of Event-to-Video Reconstruction with Implicit Neural Representations}

\titlerunning{EvINR}

\author{Zipeng~Wang\inst{1}\orcidlink{0009-0004-6541-9117} \and
Yunfan~Lu\inst{1}\orcidlink{0000-0002-7371-3189} \and
Lin~Wang\thanks{Corresponding author}\inst{1,2}\orcidlink{0000-0002-7485-4493}}

\authorrunning{Z.~Wang et al.}

\institute{Artificial Intelligence Thrust, HKUST(GZ) \and Dept. of Computer Science and Engineering, HKUST\\
\email{\{zwang253,ylu066\}@connect.hkust-gz.edu.cn, linwang@ust.hk}}

\maketitle

\begin{abstract}

Reconstructing intensity frames from event data while maintaining high temporal resolution and dynamic range is crucial for bridging the gap between event-based and frame-based computer vision. 
Previous approaches have depended on supervised learning on synthetic data, which lacks interpretability and risk over-fitting to the setting of the event simulator. 
Recently, self-supervised learning (SSL) based methods, which primarily utilize per-frame optical flow to estimate intensity via photometric constancy,  has been actively investigated. However, they are vulnerable to errors in the case of inaccurate optical flow.
This paper proposes a novel SSL event-to-video reconstruction approach, dubbed \textbf{EvINR}, which eliminates the need for labeled data or optical flow estimation.
Our core idea is to reconstruct intensity frames by directly addressing the event generation model, essentially a partial differential equation (PDE) that describes how events are generated based on the time-varying brightness signals.
Specifically, we utilize an implicit neural representation (INR), which takes in spatiotemporal coordinate $(x, y, t)$ and predicts intensity values, to represent the solution of the event generation equation. 
The INR, parameterized as a fully-connected Multi-layer Perceptron (MLP), can be optimized with its temporal derivatives supervised by events.
To make EvINR feasible for online requisites, we propose several acceleration techniques that substantially expedite the training process. 
Comprehensive experiments demonstrate that our EvINR surpasses previous SSL methods by \textbf{38}\% \textit{w.r.t.} Mean Squared Error (MSE) and is comparable or superior to SoTA supervised methods.
Project page: \url{https://vlislab22.github.io/EvINR/}.

\end{abstract}    
\section{Introduction}
\begin{figure}[t!]
    \centering
    \includegraphics[width=.9\columnwidth]{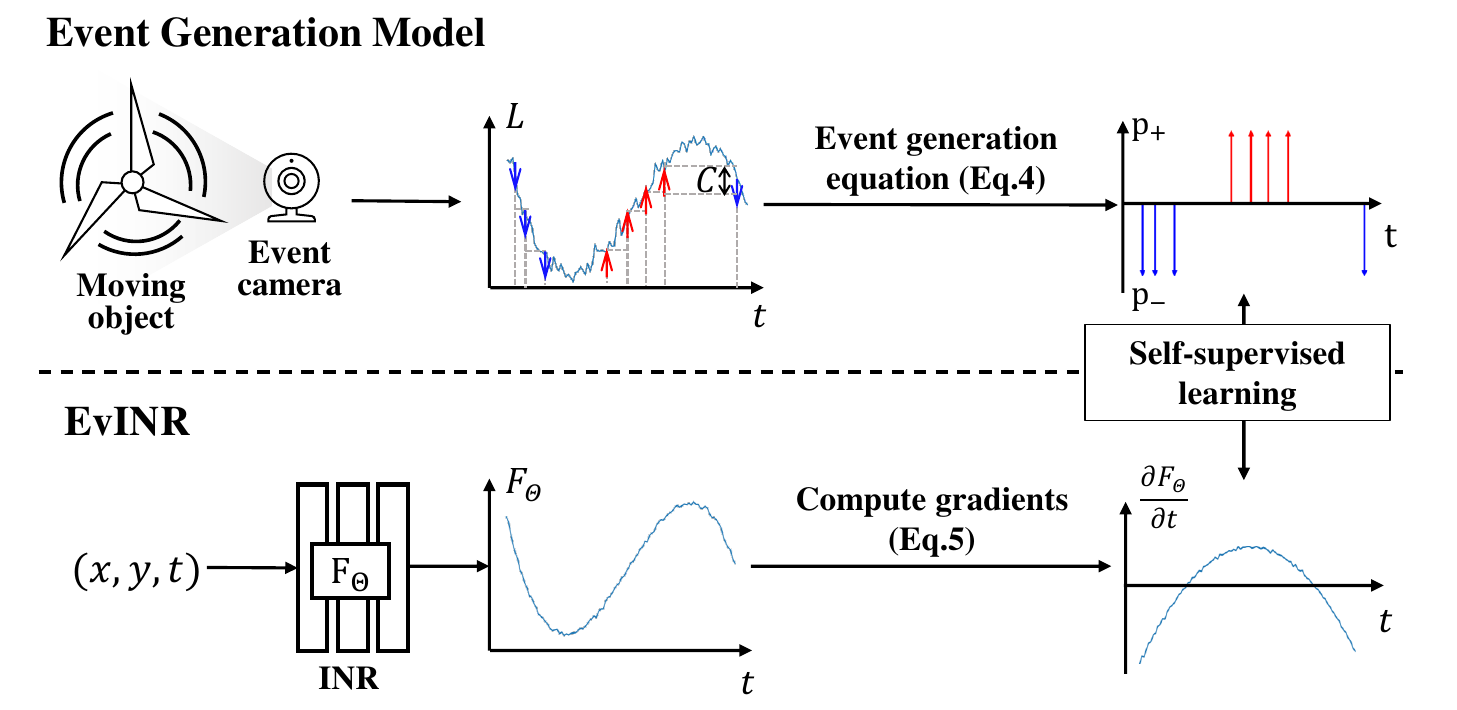}
    \caption{
    \textbf{Connection between event generation model and EvINR:} 
    Event generation model reveals the relation between discrete events and continuous temporal intensity changes, described as the event generation equation (Eq.~\ref{eq:4}).
    EvINR utilizes an INR to solve Eq.~\ref{eq:4} and recovering a continuous function of intensity \wrt time, implicitly parameterized with a fully connected MLP.
    }
    \label{figure1}
\end{figure}

Event cameras \cite{gallego2020survey, yang2015DVS} are novel sensors that offer numerous advantages over traditional frame-based cameras, including low power consumption, high dynamic range (HDR), and high temporal resolution \cite{pini2018learn}. However, their unique imaging paradigm presents a challenge when applying vision algorithms designed for frame-based cameras. To address this challenge and bridge the gap between event-based and standard computer vision~\cite{he2016resnet, he2017mask, redmon2016yolo, girshick2015fast, dosovitskiy2020vit}, many methods have been proposed to reconstruct intensity frames from events.

Early methods primarily rely on hand-crafted integrators or filters, resulting in significant loss of detail in reconstructed results \cite{bardow2016simultaneous, kim2016real, munda2018MR, scheerlinck2018CT, zhang2021whylearnit}. More recently, deep learning-based methods \cite{cadena2021spade_e2vid, rebecq2019e2vid, scheerlinck2020firenet, stoffregen2020e2vid+, weng2021etnet, zhu2022spiking} have emerged, typically supervised by large-scale synthetic datasets generated using an event simulator \cite{Gehrig2020vid2e, hu2021v2e}. However, the interpretability of these methods is limited due to the `black-box' nature of the neural networks. Furthermore, the performance of these methods is constrained by the domain gap between synthetic and real-world data \cite{stoffregen2020e2vid+}, leading to suboptimal results if the event simulator's settings do not accurately match those of the inference data.
To address these issues, some researchers have explored self-supervised learning (SSL) frameworks \cite{paredes2021ssl_e2vid, zhang2023formulating}, aiming to eliminate the dependency on labeled synthetic data. Nonetheless, these approaches still depend on event-based optical flow estimation, which is prone to over-fitting, occlusions, and non-convergence issues \cite{shiba2022event_collapse, shiba2022secrets}. As a result, the reconstructed frames often suffer from the loss of textural details and various artifacts.

To address these challenges, we revisit the event generation model \cite{gallego2020survey},  which forms a fundamental link between events and intensity.
The event generation model can be expressed as a partial differential equation (PDE), describing how events are triggered by logarithmic intensity changes that exceed a certain threshold. 
This PDE, known as the event generation equation, establishes a direct connection between discrete events and the partial derivative of the continuous intensity function \wrt time.
Our key insight is that, \textit{solving the event generation equation offers an ideal self-supervised solution for event-to-video reconstruction, thereby eliminating the need for synthetic data or optical flow estimation}. 
However, several properties of event data make solving the equation nontrivial:
\textbf{1}) events can be triggered at extremely high frequencies (up to $10^8/s$), which results in a large volume of data and a heavy computational burden.
\textbf{2}) events are inherently noisy, particularly in extreme visual conditions \cite{wang2019evgait}, which pose challenge to the robustness of solvers.
\textbf{3}) events do not capture the initial intensity, which makes determining the boundary values challenging.

Recently, implicit neural representations (INRs) have gained popularity for solving the inverse problems in 3D reconstruction or image super-resolution~\cite{chen2021liif, mildenhall2021nerf, sitzmann2020siren} by parameterizing complex signals via deep neural networks.
In this work, we find that INRs possess several key advantages that render them particularly suitable for solving the event generation equation: \textit{\textbf{they inherently accommodate a large volume of event data, exhibit high noise tolerance, and are flexible to add additional loss terms to regulate initial value}.}

In light of this, we propose a novel SSL framework, termed \textbf{EvINR}, that employs an INR to represent the solution of the event generation equation. 
Our EvINR can be directly optimized by minimizing the residual between its temporal derivatives and local intensity changes estimated from event data based on the event generation model (Sec.~\ref{3.3.1}).
Moreover, we incorporate a spatial regularization term that regularizes the relative values of adjacent pixels by constraining the magnitude of spatial gradients, which effectively reduces noise in the reconstruction process (Sec.~\ref{3.3.2}).
Although the basic implementation of EvINR yields acceptable results, its convergence on seconds-long event sequences takes minutes, limiting its real-world applicability, especially in online scenarios. To expedite EvINR's training process, we introduce several acceleration techniques, including frame-based optimization, coarse-to-fine training, and model ensembling. 
These approaches reduce the training time from minutes to seconds while not compromising the reconstruction quality (Sec.~\ref{3.4}). 

Moreover, most approaches are typically evaluated on event datasets captured by DAVIS sensors \cite{mueggler2017IJRR, stoffregen2020e2vid+, scheerlinck2019ced, zhu2018MVSEC}, making it difficult to evaluate their stability and robustness to other types of event camera~\cite{alpsentek, othereventcamera2, ryu2019othereventcamera1}. For this reason, we collected a new event dataset using an ALPIX-Eiger event camera~\cite{alpsentek}, with well-aligned events and intensity frames.

In summary, our paper makes three key contributions: 
\textbf{(I)} We propose EvINR, a concise SSL framework that solves the event generation equation via implicit neural representations for event-to-video reconstruction. 
\textbf{(II)} Our EvINR substantially outperforms previous SSL methods~\cite{paredes2021ssl_e2vid, scheerlinck2018CT, zhang2021whylearnit} and attains comparable, or even superior, performance compared to state-of-the-art supervised methods~\cite{rebecq2019e2vid, stoffregen2020e2vid+, weng2021etnet}. 
\textbf{(III)} We collect a real-world dataset with an ALPIX-Eiger event camera, complementary to the datasets captured by DAVIS cameras.

\section{Related Works}
\noindent \textbf{Implicit Neural Representations (INRs)}
have emerged as a powerful tool for parameterizing signals, such as images, videos, and audio, in a continuous manner using neural networks \cite{sitzmann2020siren}. Compared to traditional discrete signal representations, INRs offer the advantage of being able to be sampled at arbitrary resolutions with fixed memory requirements. As a result, INRs have found widespread applications in various fields, including 3D scene representation \cite{park2019deepsdf, mescheder2019occupancy, mildenhall2021nerf}, video representation\cite{chen2021nerv}, generative models \cite{skorokhodov2021adversarialINR, Schwarz2020graf, niemeyer2021giraffe}, and model compression \cite{strumpler2022Compression}.
A unique property of INRs is that they can be effectively learned from the derivatives of signals, such as the normals of 3D shapes \cite{oechsle2021unisurf, sitzmann2020siren,zhang2021nerfactor} and the gradients of images \cite{sitzmann2020siren}.
Such a characteristic has motivated us to approach the task of event-to-video reconstruction by optimizing the INR of video from its temporal derivatives.
Previous research has investigated the potential of INR for novel view synthesis using event data~\cite{klenk2023enerf, hwang2022evnerf, rudnev2022eventnerf}. These approaches reconstruct 3D neural radiance fields using multiple event sequences with known camera poses from a stationary scene. 
Most similar to our work, E-CIR~\cite{song2022cir} uses polynomial to regress the intensity function, which may only represent a short time interval (e.g., the exposure time of an image). In contrast, our method employs INR, which can represent longer time of intensity change.

\begin{figure*}[t!]
    \centering
    \includegraphics[width=0.99\linewidth]{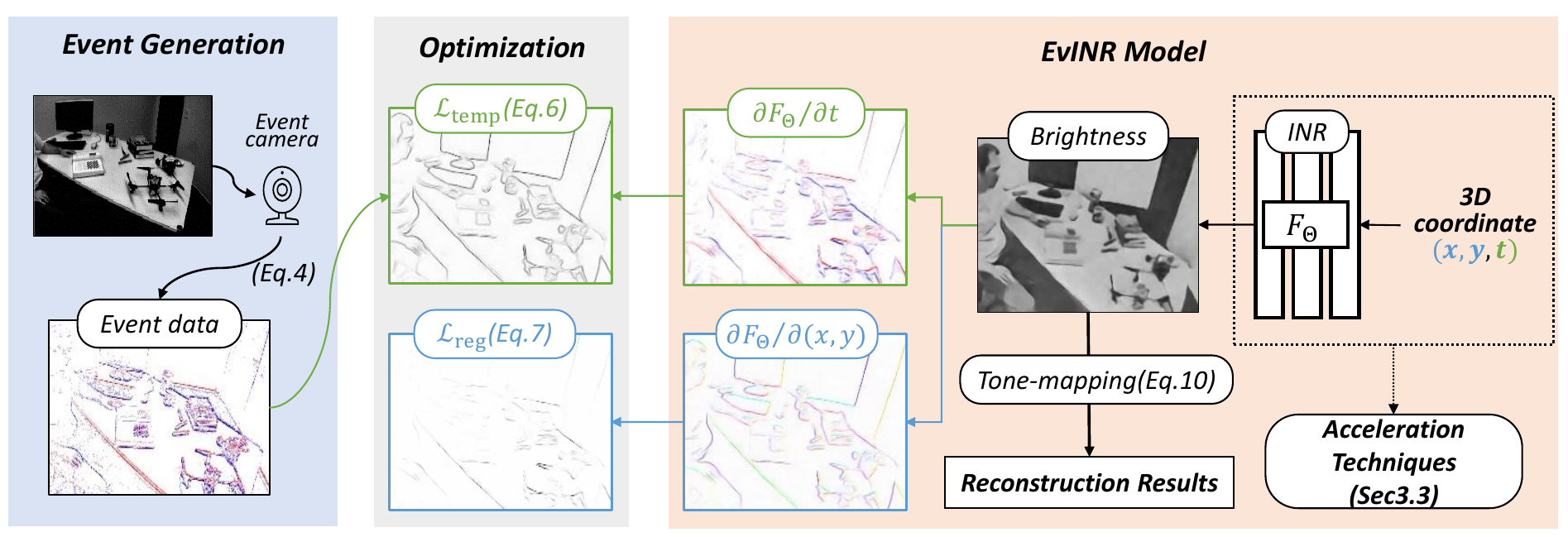}
    \caption{\textbf{Overview of EvINR}. A fully connected MLP is used to implicitly solve the event generation equation. The temporal gradient of the MLP is supervised by temporal intensity changes of events, and the spatial gradient is penalized to reduce noise.}
    \label{fig:framework}
\end{figure*}

\noindent \textbf{Event-based Video Reconstruction}
has been a hot topic in the literature. Early attempts \cite{kim2008simultaneous, cook2011interacting, bardow2016simultaneous} address this problem based on the photometric constancy, which describes the relationship between intensity gradients and optical flow. Other approaches \cite{munda2018MR, scheerlinck2018CT} are based on direct event integration without estimating optical flow.
Rebecq \etal \cite{rebecq2019e2vid} developed the first DL-based framework, called E2VID, to reconstruct intensity frames from events in an end-to-end manner, outperforming earlier techniques by a significant margin. E2VID was updated by some following research, trying to tackle problems of inference speed \cite{scheerlinck2020firenet}, the cold start problem \cite{cadena2021spade_e2vid}, and training strategy \cite{stoffregen2020e2vid+}. These methods are supervised and learned using the synthetic dataset obtained via the event simulators \cite{Gehrig2020vid2e, hu2021v2e} due to the lack of real-world datasets with well-aligned events and intensity frames. Therefore, the generalization capability of these methods is limited by the simulator-to-real gap \cite{stoffregen2020e2vid+}.

To address this problem, Federico \etal \cite{paredes2021ssl_e2vid} proposed an SSL framework with a network to estimate optical flow and another network to reconstruct intensity frames based on the photometric constancy. Zhang \etal \cite{zhang2023formulating} updated this idea into a linear inverse problem that can be solved using modern linear solvers without using deep learning. However, these methods either assume optical flow is known or estimate the optical flow using contrast maximization~\cite{zhu2018evflow, zhu2019unsupervisedevflow}. Consequently, their performance cannot be guaranteed unless high-quality optical flow can be obtained.
\textit{In contrast, our approach is based solely on the physical event generation model, leading to a more straightforward solution that also demonstrates significantly improved performance and greater flexibility.}
\section{Method}
\noindent \textbf{Overview:}
Our objective is to reconstruct intensity frames from events in a self-supervised manner, without the need for end-to-end training or optical flow estimation.
To accomplish this, we have reformulated event-based video reconstruction as solving the event generation equation. 
We employ an INR that is supervised solely by event data to represent the intensity function. An overview of our approach can be seen in Fig.\ref{fig:framework}. 
In Sec.~\ref{3.1}, we explain the event generation model and its connection to event-based video reconstruction. 
In Sec.~\ref{3.2}, we detail how to train an EvINR by supervising its spatial and temporal gradients. 
Techniques to speed up the training of EvINR for online applications are discussed in Sec.~\ref{3.3}. 
Finally, in Sec.~\ref{3.4}, we describe our collected dataset using the ALPIX event camera~\cite{alpsentek}.

\subsection{Preliminary: Event Generation Model} \label{3.1}
We begin by providing a brief overview of the event generation model~\cite{gallego2020survey}, which forms the theoretical foundation for our approach. Let $I(x,y,t)$ denote the intensity of the spatial location $(x,y)$ at time $t$ in a video. Since event cameras operate with logarithmic intensity, we denote $L(x,y,t)=\log I(x,y,t)$.

An event camera comprises a frame of independent pixels that respond to changes in the logarithmic intensity signal and produce sequences of sparse and asynchronous events. 
An event $(x_i, y_i, t_i, p_i)$ is triggered when the logarithmic intensity change surpasses a threshold $C$ since the previous event was triggered at the same pixel.
where $(x,y)$ is the spatial location of the pixel, $t$ is the timestamp, and $p \in \{-1, 1\}$ is the polarity of the logarithmic intensity change.

For simplicity, let us consider the temporal changes in logarithmic intensity of a single pixel with fixed spatial coordinates and disregard the spatial terms $(x,y)$. We can describe an event $e_i$ on that pixel using the Dirac delta function as follows:
\begin{equation}
e_i(t)= p_i \cdot C \cdot \delta(t-t_i).
\label{eq:1}
\end{equation}
Therefore, the logarithmic intensity increment $\Delta L = L(t_2) - L(t_1)$ in a time interval $[t_1, t_2]$ can be represented by the accumulation of events, which can be expressed as:
\begin{equation}
\Delta L = \int_{t_1}^{t_2} \sum_{i} e_i(t) \mathrm{d}t.
\label{eq:2}
\end{equation}
Assuming a short time interval and ignoring noise, we can approximate the logarithmic intensity increment by its first-order temporal derivative using Taylor expansion:
\begin{equation}
\frac{\Delta L}{t_2-t_1} = \frac{\partial L((t_1+t_2)/2)}{\partial t}.
\label{eq:3}
\end{equation}
By substituting Eq.~\ref{eq:2} into Eq.~\ref{eq:3}, we derive \textbf{the event generation equation} (Eq.~\ref{eq:4}), which bridges the discrete event data with the continuous temporal derivatives of logarithmic intensity: 
\begin{equation} 
\frac{\partial L((t_1+t_2)/2)}{\partial t} = \frac{1}{t_2-t_1} \int_{t_1}^{t_2} \sum_{i} e_i(t) \mathrm{d}t. \label{eq:4} 
\end{equation} 
The left-hand side of Eq.~\ref{eq:4} represents the partial derivative of logarithmic intensity with respect to time, while the right-hand side can be calculated by accumulating events. Intuitively, event-based video reconstruction can be formulated as solving Eq.~\ref{eq:4}, \ie, finding a logarithmic intensity function that satisfies the equation. 
However, as stated earlier, solving Eq.~\ref{eq:4} can be non-trivial due to the large amount, high noise rate of event data, and unknown boundary intensity values.
To address these challenges, we employ INRs to solve Eq.~\ref{eq:4} as they naturally scale up to large data and have high noise tolerance~\cite{hwang2022evnerf}. Boundary intensity values can also easily be regularized by injecting natural image priors as additional loss functions.

\subsection{Learning INRs from Events} \label{3.2}
We aim to learn an INR $F_\Theta$ given an event stream $\{e_i\}_{i=0}^N$. 
Here, $\Theta$ represents the parameters of a fully connected MLP that predicts the logarithmic intensity value $\hat{L}$ at any spatiotemporal coordinate $(x, y, t)$. 
We adapt the event frame representation~\cite{rebecq2019e2vid} and stack the given events uniformly into $T$ event frames.

\subsubsection{Temporal Supervision} \label{3.3.1}
The INR is optimized by directly minimizing the temporal loss between the predicted logarithmic intensity change $\Delta \hat{L}$ and the logarithmic intensity change estimated by event accumulation $\Delta L$. Here $\Delta \hat{L}$ can be derived by accumulating events as Eq.~\ref{eq:2} and $\Delta \hat{L}$ is the change of $\hat{L}$ with respect to $t$ obtained by performing double back-propagation~\cite{paszke2017automatic} of $F_\Theta$:
\begin{equation}
\Delta \hat{L} = \frac{\partial F_\Theta((t_1+t_2)/2)}{\partial t} \cdot (t_2 - t_1),
\label{eq:5}
\end{equation}
Here we adopt the mean squared error (MSE) loss for temporal supervision:
\begin{equation}
\mathcal{L}_\text{temp} = (\Delta L - \Delta \hat{L}) ^2,
\label{temporal_loss}
\end{equation}
where $\Delta L$ and $\Delta \hat{L}$ are given by Eq.~\ref{eq:2} and Eq.~\ref{eq:5}, respectively.

\subsubsection{Spatial Regularization} \label{3.3.2}
Although the temporal supervision can estimate the intensity function, the results may still contain unnatural artifacts as the INR network $F_\Theta$  has no prior knowledge of the initial intensity values of each pixel.
To address this issue, we introduce a spatial regularization term that encourages the solution to be in the space of natural-looking images. 
We adopt Tikhonov regularization \cite{bishop2006pattern} that penalizes the spatial gradients of logarithmic intensity:
\begin{equation}
\mathcal{L}_{\text{reg}} = (\frac{\partial F_\Theta}{\partial x})^2 + (\frac{\partial F_\Theta}{\partial y})^2.
\label{eq:9}
\end{equation}

It is worth noting that we do not employ more complex regularization methods, such as CNN denoisers \cite{zhang2021plug} as used in \cite{zhang2023formulating}, because they typically utilize large network models that can significantly slow down the training process.

\subsubsection{Optimizing EvINR} \label{3.3.3}
The overall objective is given by the equation:
\begin{equation}
\mathcal{L} = \mathcal{L}_{\text{temp}} + \lambda \mathcal{L}_{\text{reg}},
\end{equation}
where $\mathcal{L}_{\text{temp}}$ and $\mathcal{L}_{\text{reg}}$ were introduced in Sec.~\ref{3.3.1} and Sec.~\ref{3.3.2}, respectively, as the temporal supervision and spatial regularization term.  $\lambda$ is a hyper-parameter used to adjust the weight of the spatial regularization term.

\subsubsection{Tone-mapping} \label{3.3.4}
The output of the EvINR is the logarithmic intensity of the reconstructed frames. We first use the exponential function to convert the predicted logarithmic intensity values to high dynamic range (HDR) intensity values $I \in [0, \infty)$:
\begin{equation}
I(x, y, t) = \text{exp}(F_\Theta(x, y, t)).
\end{equation}
Then, we adopt the Reinhard function \cite{reinhard2002} to map the HDR intensity values into the low dynamic range $[0,1]$:
\begin{equation}
\Gamma(I) = (\frac{I}{I+1})^{\gamma}.
\label{eq:12}
\end{equation}
where $\gamma$ is a hyper-parameter used to control the contrast of $\Gamma(I)$. We assess our reconstruction performance by applying $\Gamma(I)$ in all experiments.

\subsection{Accelerating EvINR} \label{3.3}
\begin{figure*}[t!]
    \centering
    \includegraphics[width=0.88\linewidth]{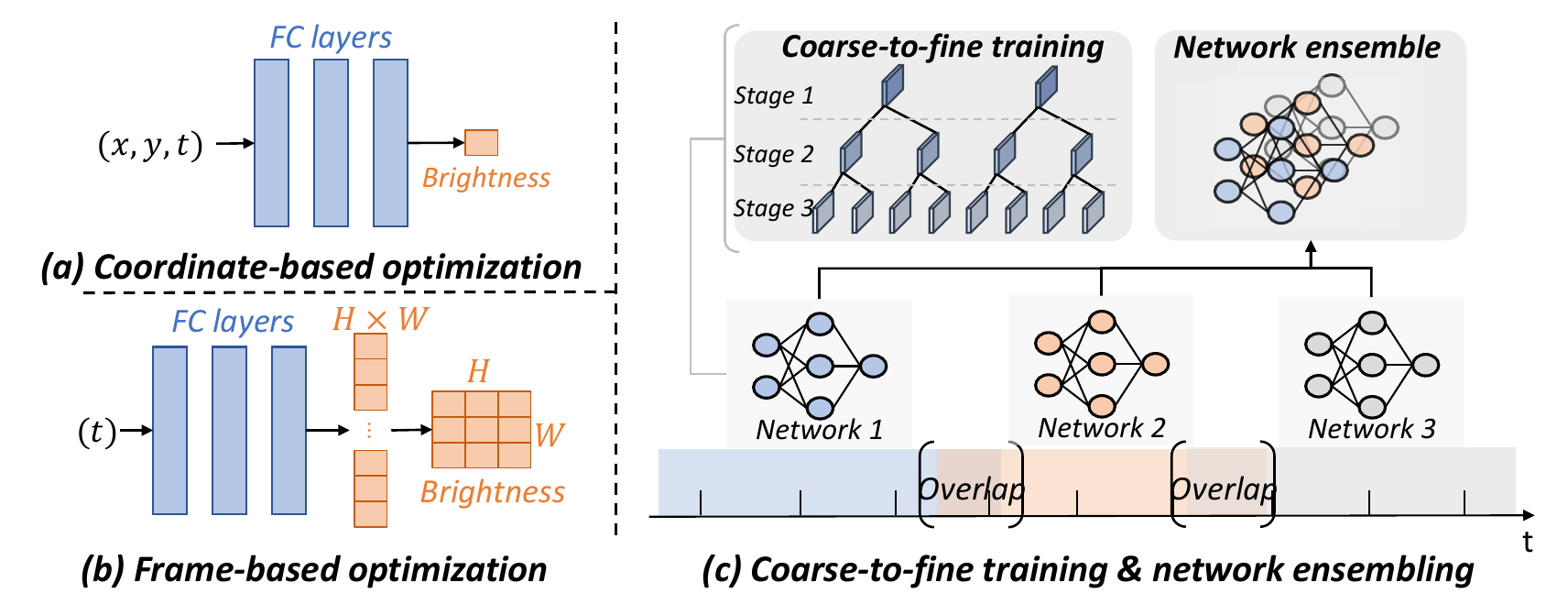}
    \caption{\textbf{Overview of acceleration techniques}. (a) and (b) illustrate the difference between the basic coordinate-based and our frame-based optimization, respectively. (c) depicts our proposed coarse-to-fine training scheme and network ensembling technique.}
    \label{fig:acceleration}
\end{figure*}
While solving the event generation equation using the basic implementation of EvINR, as described in Sec.~\ref{3.3}, yields satisfactory reconstruction results, it is not efficient enough for online tasks. Optimizing one INR network takes about 20 minutes, and a network can only represent approximately 1 second of a sequence. Reducing the training time or increasing the sequence time leads to severe performance degradation. To enable online usage for EvINR, we propose several techniques to reduce the training time and increase the representation capacity for EvINR, as illustrated in Fig.~\ref{fig:acceleration}.

\noindent\textbf{Coordinate-based to frame-based optimization: }
The basic EvINR algorithm is designed to learn a mapping from 3D coordinates $(x,y,t)$ to the logarithmic intensity $L$. However, this coordinate-based optimization requires the network to 'remember' $H\times W\times T$ coordinate-to-logarithmic intensity mappings, which can be highly complex and lead to slow convergence. 
To address this issue, we propose a frame-based optimization scheme that learns a mapping from timestamps $t$ to the logarithmic intensity of all pixels at that timestamp. The frame-based optimization scheme can be represented as:
\begin{equation}
    F_\Theta (t) = [L(i,j,t)]_{1\le i \le H, 1 \le j \le W},
    \label{eq:11}
\end{equation}
where the right hand-side represent a matrix that contains all logarithmic intensity of all pixels at $t$.
The frame-based optimization approach requires the network to remember only $T$ coordinate-to-logarithmic intensity mappings, resulting in a significant reduction in complexity and faster convergence. Empirical results show that this approach reduces the convergence time by two orders of magnitude, from minutes to seconds.
The comparison between coordinate-based and frame-based optimizations is illustrated in Fig.~\ref{fig:acceleration}~(a) and (b).

\noindent\textbf{Coarse-to-fine training: }\label{3.4.2}
We adopt a coarse-to-fine training scheme, which enables the network to learn overall logarithmic intensity changes before focusing on finer details. 
We structure the training process into $s$ distinct stages, at each of which we increase the temporal resolution by a factor of 2. Initially, we divide the event sequence into $N$ segments of equal length and optimize the EvINR in accordance with Eq.~\ref{temporal_loss} over a specified number of iterations. Subsequently, we proceed to bisect each of these $N$ segments into two smaller ones, ensuring that they contain an equal number of events. 
For all the experiments, we set the number of stages $s$ to 3 and scheduled the upsampling to occur after 100 and 200 iterations, respectively.
This approach accelerates training by reducing the number of event frames needed in the early stages by approximately 2 times.

\noindent\textbf{Network ensembling:}
As each EvINR network requires only about 1GB of GPU memory, we further leverage network ensemble techniques~\cite{hansen1990ensemble} to train $N$ EvINR networks simultaneously on a single GPU to achieve higher parallelism. This approach enables us to exploit the computational resources of modern GPUs more effectively and speed up the training process. We keep overlap periods between nearby networks to keep logarithmic intensity predictions constant among all networks.
Fig.~\ref{fig:acceleration}(c) depicts the coarse-to-fine training process and network ensembling techniques employed in our approach.

\subsection{Dataset Collection} \label{3.4}
Most event-to-video reconstruction approaches are typically evaluated on event datasets collected using DAVIS sensors \cite{brandli2014DAVIS}, such as IJRR \cite{mueggler2017IJRR}, HQF \cite{stoffregen2020e2vid+}, MVSEC \cite{zhu2018MVSEC}, and CED \cite{scheerlinck2019ced}. 
Recently, other types of event cameras~\cite{othereventcamera2, ryu2019othereventcamera1, alpsentek} have been developed, which share the same event generation model with DAVIS sensors but may differ in detailed configurations and settings.
Therefore, it is essential to verify the generalization capability of existing methods and our EvINR on those new types of event sensors.

To fill the gap, we introduce a new real-world dataset, called the \textbf{A}LPIX \textbf{E}vent \textbf{D}ataset (AED), which is collected using an ALPIX-Eiger event camera~\cite{alpsentek}, featuring static scenes accompanied by gradual camera motions. The camera provides well-aligned RGB frames and color events. The RGB frames have a resolution of $3264\times 2448$ and the events have a resolution of $1632\times 1224$. The AED dataset includes seven video sequences with diverse scenes, such as streets, buildings, indoor scenes, textures, and tools, and each approximately lasts for ten seconds.
Note that, in this paper, we only focus on reconstructing grayscale frames to keep consistency with previous works~\cite{paredes2021ssl_e2vid, zhang2023formulating} and also for a fair comparison. Therefore, we first demosaic the RGB frames and events according to the Quad Bayer pattern, resulting in intensity frames with a resolution of $816\times 612$ and event data with a resolution of $408\times 306$. Details on the post-processing of AED dataset can be found in the supplementary material.

\section{Experiments}

\begin{figure*}[t!]
    \centering
    \includegraphics[width=1\linewidth]{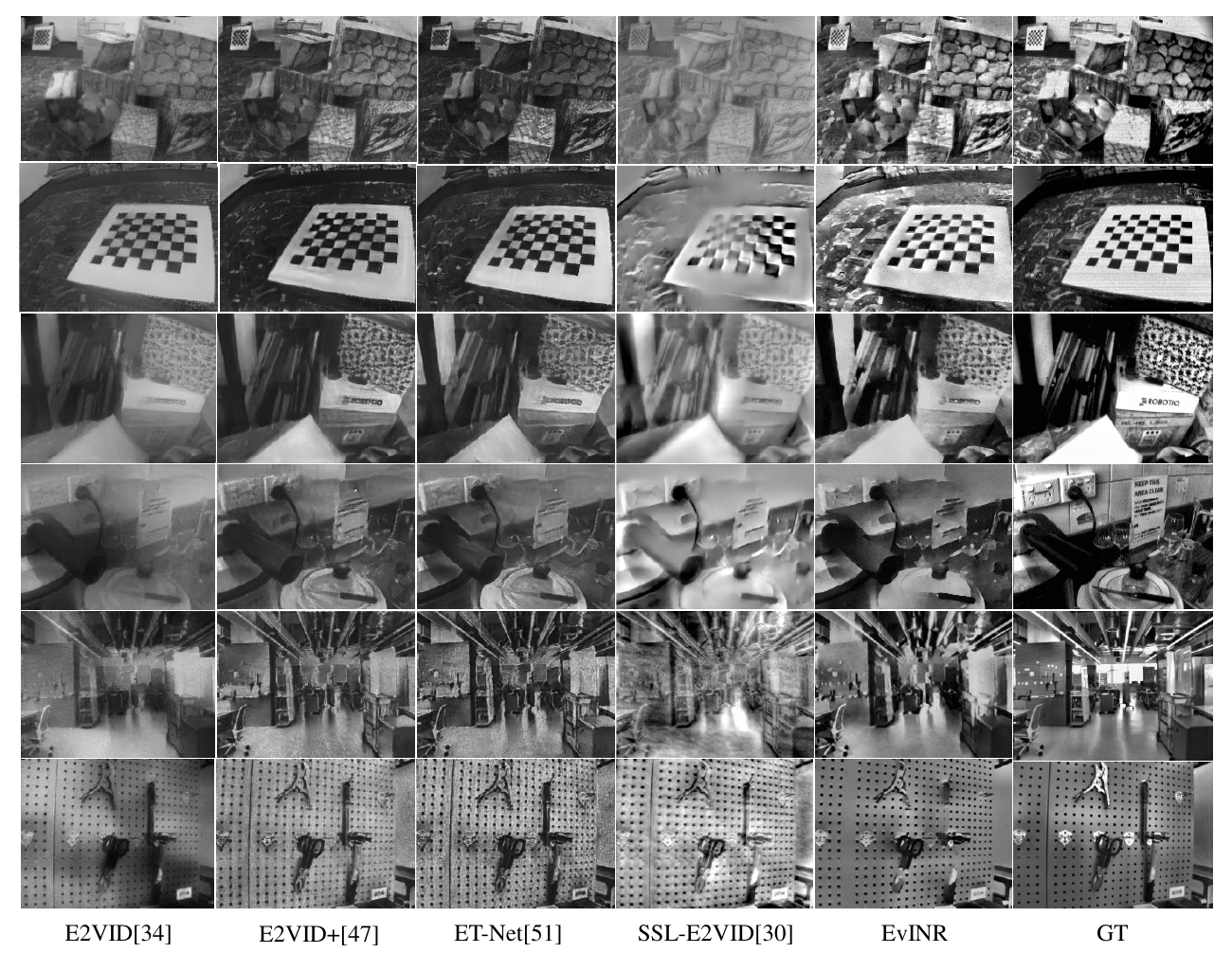}
    \caption{Qualitative comparison with baseline methods on IJRR(Row 1\&2), HQF (Row 3\&4) and AED(Row 5\&6).}
    \label{fig:comparison}
\end{figure*}

\subsection{Experiments Settings}
\noindent\textbf{Datasets:} 
We conduct experiments to evaluate the effectiveness of our proposed method using three datasets: IJRR \cite{mueggler2017IJRR}, HQF \cite{stoffregen2020e2vid+}, and our AED dataset. The IJRR dataset consists of intensity frames and events in 25 real scenes and 2 synthetic scenes, captured by a DAVIS240C camera \cite{brandli2014DAVIS}.
HQF dataset provides 14 event data sequences captured with two DAVIS240C cameras, delivering well-exposed and clear intensity frames. The spatial resolution of both the IJRR and HQF datasets is $240\times180$. The AED dataset contains 7 event sequences with a resolution of $408\times306$. 
More details of the exact time split can be found in the supplementary material.

\noindent\textbf{Evaluation Metrics:}
We evaluate the efficacy of our method using several image quality metrics, including mean squared error (MSE), structural similarity (SSIM) \cite{wang2004SSIM}, and learned perceptual image patch similarity (LPIPS) \cite{zhang2018LPIPS}. 

\noindent\textbf{Implementation Details:}
We use a SIREN network \cite{sitzmann2020siren} with three hidden layers and 512 neurons per layer as our INR network. 
We partition each event sequence into sub-sequences lasting 5 seconds each and concurrently train all sub-sequences using model ensembling for 300 iterations.
We first stack the events within $\frac{1}{32}$ second into an event frame and temporal upsample the event frames in 100 and 200 iterations, as described in Sec.~\ref{3.4.2}.
We adopt the Adam optimizer \cite{kingma2014adam} with a learning rate of $1\mathrm{e}{-4}$ and exponentially reduce the learning rate every 10 iterations with a decay rate of 0.95. The weight of spatial regularization $\lambda$ is set to 0.05, and $\gamma$ in Eq.~\ref{eq:12} is set to 0.6. The activated threshold $C$ is set to 1 for the IJRR and HQF datasets and 0.25 for the AED dataset.  
The training process takes approximately 8 seconds on a single RTX3090 GPU.

\subsection{Evaluation of Video Reconstruction}
We assess the effectiveness of our approach by comparing it against eight state-of-the-art (SoTA) methods, classified based on the amount of data required for training. 
The methods are categorized into supervised learning (SL) methods that utilize synthetic ground-truth intensity frames for supervision and self-supervised learning (SSL) methods that rely solely on event data. 
Specifically, we compare against:
\textbf{1)} Five SL methods: FireNet \cite{scheerlinck2020firenet}, E2VID \cite{rebecq2019e2vid}, FireNet+ \cite{stoffregen2020e2vid+}, E2VID+ \cite{stoffregen2020e2vid+}, and ET-Net \cite{weng2021etnet}.
\textbf{2)} Three SSL method: SSL-E2VID \cite{paredes2021ssl_e2vid}, HF \cite{scheerlinck2018CT} and ELRP~\cite{zhang2023formulating}.
Reconstructed results for all methods were generated at each timestamp of the intensity frame. 
We use optical flow estimation from the FlowNet of SSL-E2VID~\cite{paredes2021ssl_e2vid} for ELRP~\cite{zhang2023formulating}.
We apply Contrast Limited Adaptive Histogram Equalization (CLAHE) \cite{yadav2014CLAHE} to both ground truth and synthesized frames before evaluation, following \cite{paredes2021ssl_e2vid}. 
Note that the quantitative results for FireNet \cite{scheerlinck2020firenet}, E2VID \cite{rebecq2019e2vid}, FireNet+ \cite{stoffregen2020e2vid+}, E2VID+ \cite{stoffregen2020e2vid+}, and ET-Net \cite{weng2021etnet} are referred from a recent benchmark paper~\cite{ercan2023evreal}. For the visual comparison, we obtained the visualization results by downloading the publicly available checkpoints and test them in the local environment.
\begin{table*}
\centering
\caption{Comparison of quantitative results on the IJRR, HQF, and AED datasets. Bold values indicate the best results among all methods, while underlined values indicate the best results among SSL methods.}
\resizebox{0.9\linewidth}{!}{%
\begin{tabular}{ccccccccccc}
\toprule[1pt]
\multirow{2}{*}{}   & \multirow{2}{*}{Methods} & \multicolumn{3}{c}{IJRR} & \multicolumn{3}{c}{HQF} & \multicolumn{3}{c}{AED} \\ \cline{3-11} 
                    &                          & MSE    & SSIM   & LPIPS  & MSE    & SSIM   & LPIPS & MSE     & SSIM   & LPIPS  \\ \hline
\multirow{5}{*}{SL} & FireNet\cite{scheerlinck2020firenet}                  & 0.131  & 0.502  & 0.320  & 0.094  & 0.533  & 0.441 & 0.074   & 0.298  & 0.579  \\
                    & E2VID\cite{rebecq2019e2vid}                    & 0.212  & 0.424  & 0.350  & 0.127  & 0.540  & 0.382 & \textbf{0.056}   & 0.424  & 0.500  \\
                    & FireNet+\cite{stoffregen2020e2vid+}                 & 0.063  & 0.555  & 0.290  & 0.040  & 0.614  & 0.314 & 0.094   & 0.232  & 0.566  \\
                    & E2VID+\cite{stoffregen2020e2vid+}           & 0.070  & 0.560  & 0.236  & 0.036  & 0.643  & 0.252 & 0.074   & 0.345  & 0.462  \\
                    & ET-Net\cite{weng2021etnet}                   & 0.047  & 0.617  & \textbf{0.224}  & \textbf{0.032}  & \textbf{0.658}  & \textbf{0.260} & 0.084   & 0.312  & 0.482  \\ \hline
\multirow{4}{*}{SSL}& SSL\_E2VID\cite{paredes2021ssl_e2vid}               & 0.097  & 0.473  & 0.409  & 0.070  & 0.480  & 0.464 & 0.094   & 0.316  & 0.453  \\ 
                    & HF\cite{scheerlinck2018CT}                       & 0.164  & 0.334  & 0.658  & 0.133  & 0.232  & 0.670 & 0.080   & 0.240  & 0.943  \\
                    & ELRP\cite{zhang2023formulating}             & 0.080  & 0.437  & 0.485  &0.074 &0.450 &0.474 &0.084 &0.305 &0.473
                    \\ \hdashline
                    & Ours                     & \textbf{\underline{0.047}}  & \textbf{\underline{0.628}}  & \underline{0.251}  & \underline{0.048}  & \underline{0.531}  & \underline{0.333} & \underline{0.067}   & \textbf{\underline{0.458}}  & \textbf{\underline{0.366}}  \\ \bottomrule[1pt]
\end{tabular}%
}
\label{tab:comparison}
\end{table*}

The quantitative results are presented in Table \ref{tab:comparison}. Our method demonstrates superior performance compared to the best SL methods, with improvements of 7\%, 13\%, and 4\% in terms of MSE, SSIM, and LPIPS respectively on the IJRR dataset~\cite{mueggler2017IJRR}. Although the gap between our method and SL methods widens on the HQF dataset~\cite{stoffregen2020e2vid+} due to the lower event density, our method remains comparable. On the AED dataset, our method outperforms the state-of-the-art SL methods, E2VID+ and ET-Net, by a clear margin for all three metrics. Notably, our method shows a significant improvement over previous SSL methods. In particular, compared with SSL-E2VID, it improves MSE, SSIM, and LPIPS by 35\%, 25\%, and 21\%, respectively. 

A qualitative comparison of our method with the baseline methods is depicted in Fig.~\ref{fig:comparison}. Our method produces intensity frames with better contrast and overall visual quality, compared with other methods that suffer from issues such as foggy effects, artifacts, and loss of detailed structures. It is worth noting that the performance of E2VID+ and ET-Net degrades significantly on the AED dataset, \textit{which suggests that their training strategy may overfit the DAVIS sensor setting}. Additional results can be found in the supplementary material.

\subsection{Ablation Study}
\begin{figure}[tp]
    \centering
    \includegraphics[width=0.9\linewidth]{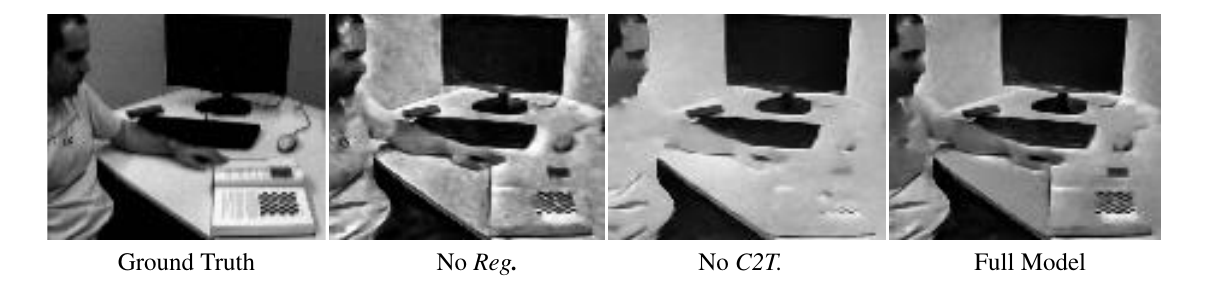}
    \caption{The impact of removing the spatial regularization and coarse-to-fine training.}
    \label{fig:ablation}
\end{figure}

\begin{table}[tp]
\centering
\caption{Ablation of the spatial regularization term (\textit{Reg.}) and coarse-to-fine training (\textit{C2F.}) on the IJRR~\cite{mueggler2017IJRR} dataset.}
\resizebox{0.45\columnwidth}{!}{
\begin{tabular}{lccccc}
\toprule[1pt]
\multicolumn{1}{c}{} & MSE   & SSIM  & LPIPS & Time(s) & FPS\\ \hline
\textit{Base}        & 0.063 & 0.542 & 0.357 & 17.29 & 13.77   \\
\textit{Base+Reg.}   & 0.044 & 0.612 & 0.275 & 17.44 & 13.64   \\
\textit{Base+C2F.}   & 0.061 & 0.566 & 0.343  & \textbf{7.94}  & \textbf{31.33}    \\                      \hdashline
Full Model           & \textbf{0.047} & \textbf{0.658} & \textbf{0.251} & 8.08 & 29.45\\ 
\bottomrule[1pt]
\end{tabular}
}
\label{tab:ablation}
\end{table}
We conduct ablation experiments on the IJRR dataset to assess the significance of each individual component of our method, and the results are presented in Tab.~\ref{tab:ablation}. 
Our method's basic implementation, which doesn't include spatial regularization term as described in Sec.~\ref{3.3.1} or use coarse-to-fine training as described in Sec.~\ref{3.4.2}, is called \textit{Base}. 
We also evaluate the performance of two modified versions of the full model: \textit{Base + Reg.}, which adds a spatial regularization term (Eq.~\ref{eq:9}) with $\lambda=0.05$, and \textit{Base + C2F}, which uses coarse-to-fine training. 
Experimental results show that both spatial regularization and coarse-to-fine training contribute to the improved performance of our full model. Fig.~\ref{fig:ablation} confirms the effectiveness of these two modules, showing that removing the spatial regularization term introduces noticeable noise (e.g., in the background), while removing the coarse-to-fine training scheme leads to an increase in training time and missing details (e.g., in the people's clothes and the book).

\subsection{Discussions}
\noindent{\textbf{Training speed:}}
\begin{table}[tp]
\centering
\caption{Impact of hyper-parameters of model ensembling on the training speed and reconstruction performance.}
\resizebox{0.4\columnwidth}{!}{%
\begin{tabular}{ll|lllll}
\toprule[1pt]
N & $\tau$  & MSE    & SSIM  & LPIPS & Time(s) & FPS   \\ \hline
1  & 50  & 0.084  & 0.46  & 0.380 & \textbf{15.71}   & \textbf{75.66} \\ 
5  & 10  & 0.0688 & 0.600 & 0.253 & 16.91   & 70.31 \\
10   & 5 & \textbf{0.0625} & \textbf{0.612} & \textbf{0.244} & 18.32   & 64.90 \\
50   & 1 & 0.0875 & 0.488 & 0.343 & 29.43   & 40.39 \\
\bottomrule[1pt]
\end{tabular}%
}
\label{tab:hyper-parameters}
\end{table}
To assess the training speed of EvINR, we conducted an analysis using various model ensembling configurations on the whole \textit{Calibration} sequence from the IJRR dataset~\cite{mueggler2017IJRR}. 
We divide the event sequence, which had a total duration of 50 seconds, into $N$ partitions, with each partition having a duration of $\tau$ seconds. Subsequently, we trained $N$ EvINR models in parallel, incorporating the model ensembling techniques outlined in Sec.~\ref{3.4}. The experimental findings are presented in Table~\ref{tab:hyper-parameters}. By training 10 EvINR models, each corresponding to 5 seconds of the event sequence, we achieved a performance gain of over 26\%, while only requiring an additional 14\% of training time compared to training a single EvINR model on the entire sequence. However, we observed a significant drop in performance when increasing the partition size to 50, suggesting that a 1-second sequence is insufficient for EvINR to converge to a stable solution of the event generation equation.
We also note that the optimal choice of hyper-parameters may differ for different event sequences due to the per-scene optimization nature of our INR approach.

\begin{figure}[htp]
    \centering
    \includegraphics[width=0.85\linewidth]{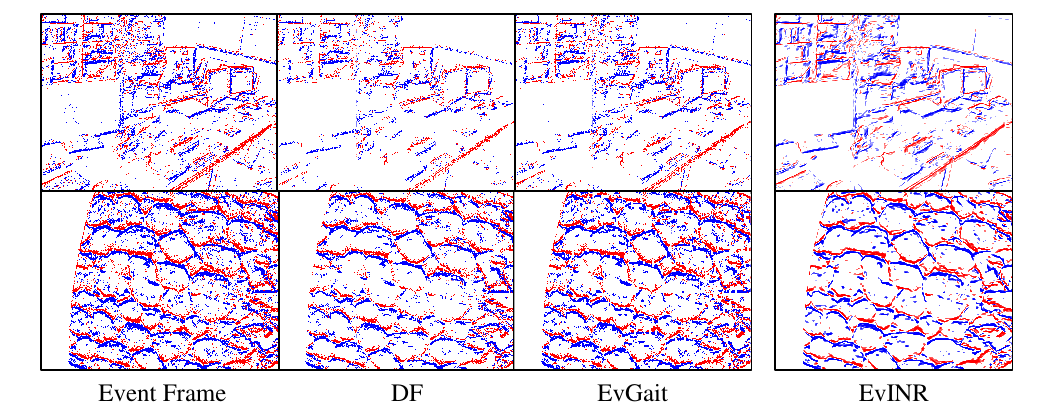}
    \caption{Qualitative comparison with baseline methods of event stream denoising.}
    \label{fig:event-enhancemnet}
\end{figure}

\noindent{\textbf{Event enhancement:}}
Our approach provides a smooth and continuous representation of event data by representing events triggered within a small time window $\Delta t$ as $\frac{\partial F_{\Theta}}{\partial t}\Delta t$. This event representation automatically reduces noise and preserves critical information through INR optimization, making it highly robust to noise. Fig.~\ref{fig:event-enhancemnet} compares the denoising results of our INR representation with several SoTA denoising methods~\cite{feng2020density, wang2019evgait}.

\begin{figure}[!h]
    \centering
    \includegraphics[width=0.9\linewidth]{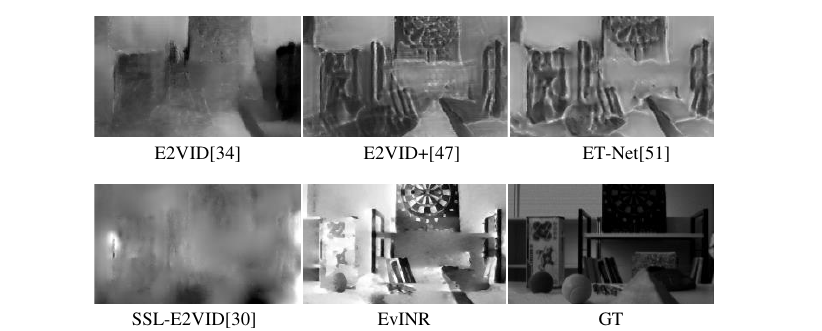}
    \caption{The reconstructed initial frame from different compared methods.}
    \label{fig:first}
\end{figure}

\noindent{\textbf{Early frames reconstruction:}}
\cite{cadena2021spade_e2vid} suggested that RNN-based techniques for video reconstruction, \eg E2VID~\cite{rebecq2019e2vid}, require an initialization period to achieve satisfactory results. Consequently, the initial frames generated by these methods may be of poor quality, restricting their usefulness to short event sequences. Conversely, our proposed approach is capable of producing realistic outcomes using a minimal number of events. Fig.~\ref{fig:first} illustrates a comparison of the first frame generated by our method and RNN-based techniques, demonstrating our ability to rapidly create high-quality outputs with minimal input. 

\noindent{\textbf{Inference speed:}}
Compared to other SoTA approaches, our method offers a significant improvement in terms of inference speed, as demonstrated in Tab.~\ref{tab:speed}. The main reason for this improvement is that other methods require converting event data into event frames or voxel grids during the inference period, which adds significant overhead. However, our method only requires the spatiotemporal coordinates as input, thus avoiding this issue.

\begin{table}[!t]
\centering
\caption{Comparison of inference time in terms of frames-per-second (fps) at two resolutions from IJRR~\cite{mueggler2017IJRR} and AED dataset respectively. 
}
\resizebox{0.5\columnwidth}{!}{%
\begin{tabular}{ccccc}
\toprule[1pt]
       Resolution      & FireNet & E2VID  & ET-Net & Ours   \\ \hline
(240,   180) & 118.24  & 100.96 & 36.1   & 178.19 \\
(408,   306) & 28.83   & 28.56  & 18.27  & 55.37  \\ 
\bottomrule[1pt]
\end{tabular}%
}
\label{tab:speed}
\end{table}

\section{Conclusion}
\vspace{-3pt}
This paper introduced EvINR, a SSL method for event-to-video reconstruction that relieves the need for synthetic data or optical flow estimation. 
We, for the first time, show that high-quality videos can be reconstructed in a self-supervised and interpretable way without time-consuming end-to-end training.
Our method is based on directly solving the event generation model via optimizing an INR whose temporal derivative is self-supervised by events and spatial derivative is regularized to reduce artifacts. 
Additionally, we propose several acceleration techniques that significantly reduce the training time of EvINR, making it applicable for online tasks.
Experiments show that our approach significantly outperforms the previous SSL methods, and is competitive with the SoTA supervised methods. Our approach also demonstrates superior interpretability and robustness to various event dataset. 
Overall, our work contributes to the advancement of event-to-video reconstruction and offers a promising direction for future research that combines INRs with event data.

\noindent \textbf{Limitations and Future Work:}
The current parameter size of EvINR takes up approximately the same amount of storage as the original event data. In future work, we plan to explore network pruning and quantization techniques to further reduce the parameter size.

\section*{Acknowledgments}
This paper is supported by the National Natural Science Foundation of China (NSF) under Grant No. NSFC22FYT45, the Guangzhou City, University and Enterprise Joint Fund under Grant No.SL2022A03J01278, and Guangzhou Fundamental and Applied Basic Research (\textbf{Grant Number}: 2024A04J4072)


%
%
\bibliographystyle{splncs04}
\bibliography{main}
\end{document}